\documentclass[10pt,twocolumn,letterpaper]{article}

\usepackage{cvpr}
\usepackage{times}
\usepackage{epsfig}
\usepackage{graphicx}
\usepackage{amsmath}
\usepackage{amssymb}

\usepackage[font=small,labelfont=bf]{caption} 
\usepackage{booktabs,siunitx}
\usepackage{tabularx}
\usepackage[usenames, dvipsnames]{xcolor}
\usepackage{multirow,bigdelim}
\usepackage{subcaption}
\usepackage{float}
\usepackage{enumitem}

\graphicspath{{figures/}}

\newcommand{\B}[1]{{\textbf{#1}}}
\newcommand{\SC}[1]{{\textsc{#1}}}
\newcommand{\commentout}[1]{}

\newcommand{\refeqn}[1]{Equation~\ref{#1}}
\newcommand{\reffig}[1]{Figure~\ref{#1}}
\newcommand{\reftbl}[1]{Table~\ref{#1}}

\definecolor{citecolor}{RGB}{34,139,34}
\usepackage[pagebackref=true,breaklinks=true,letterpaper=true,colorlinks,
citecolor=citecolor,bookmarks=false]{hyperref}

\cvprfinalcopy %

\ifcvprfinal\fi
\begin{document}

\title{Grounded Human-Object Interaction Hotspots from Video \\ (Extended Abstract)}

\author{Tushar Nagarajan\thanks{Work done during internship at Facebook AI Research.}\\
UT Austin\\
{\tt\small tushar@cs.utexas.edu}
\and
Christoph Feichtenhofer\\
Facebook AI Research\\
{\tt\small feichtenhofer@fb.com}
\and
Kristen Grauman\\
Facebook AI Research\\
{\tt\small grauman@fb.com\thanks{On leave from UT Austin (\texttt{grauman@cs.utexas.edu}).}}
}

\maketitle

\begin{abstract}
Learning how to interact with objects is an important step towards embodied visual intelligence, but existing techniques suffer from heavy supervision or sensing requirements.  
We propose an approach to learn human-object interaction ``hotspots" directly from video.
Rather than treat affordances as a manually supervised semantic segmentation task, our approach learns about interactions by watching videos of real human behavior and anticipating afforded actions.
Given a novel image or video, our model infers a spatial hotspot map indicating how an object would be manipulated in a potential interaction---even if the object is currently at rest.
Through results with both first and third person video, we show the value of grounding affordances in real human-object interactions.  Not only are our weakly supervised hotspots competitive with strongly supervised affordance methods, but they can also anticipate object interaction for novel object categories.  
Project page: \url{http://vision.cs.utexas.edu/projects/interaction-hotspots/}
\end{abstract}

\vspace*{-0.1in}
\section{Introduction}

Today's visual recognition systems know how objects \emph{look}, but not how they \emph{work}.
Understanding how objects function is fundamental to moving beyond passive perceptual systems (\eg, those trained for image recognition)  to active, embodied agents that are capable of both perceiving and interacting with their environment---whether to clear debris in a search and rescue operation, cook a meal in the kitchen, or even engage in a social event with people.
Gibson's theory of affordances~\cite{gibson1979ecological} provides a way to reason about object function.  It suggests that objects have ``action possibilities" (\eg, a chair affords sitting, a broom affords cleaning), and has been studied extensively in computer vision and robotics in the context of action, scene, and object understanding~\cite{hassanin2018visual}.

However, the abstract notion of ``what actions are possible?'' is only half the story. For example, for an agent tasked with sweeping the floor with a broom, knowing that the broom handle \emph{affords holding} and the broom \emph{affords sweeping} is not enough.  
The agent also needs to know \emph{how} to interact with different objects, including
the best way to grasp the object,
the specific points on the object that need to be manipulated for a successful interaction, %
how the object is used to achieve a goal, and even what it suggests about how to interact with \emph{other} objects. %

\begin{figure}[t!]
\centering
\includegraphics[width=\columnwidth]{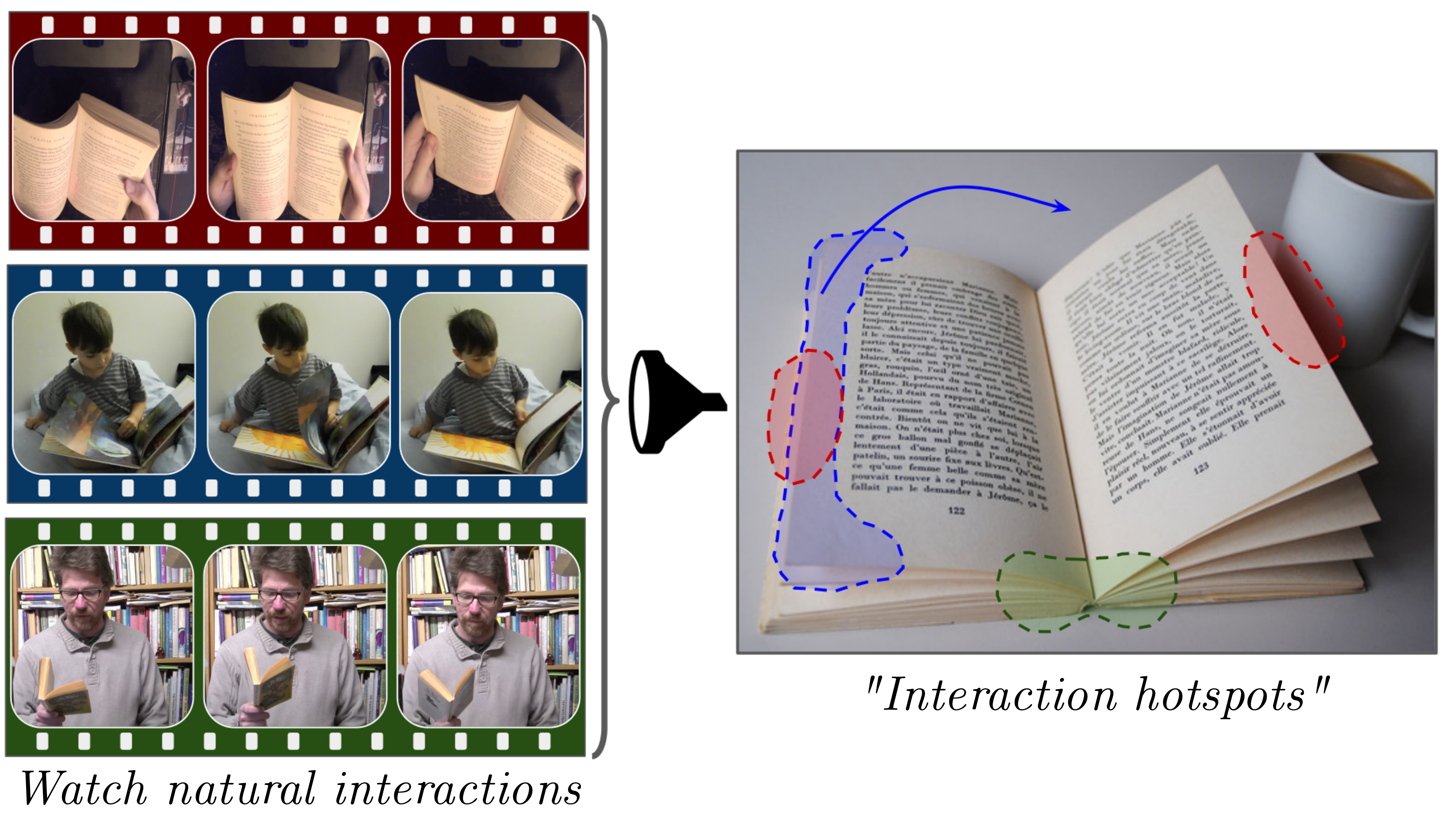}
\caption{We propose to learn object affordances directly from videos of people naturally interacting with objects.  The resulting  representation of ``interaction hotspots" is grounded in real human behavior from video, rather than manual image annotations.}
\label{fig:concept}
\end{figure}

Learning how to interact with objects is challenging.  Traditional methods face two key limitations.  First, methods that consider affordances as properties of an object's shape or appearance~\cite{myers2015affordance,grabner2011makes,hermans2011affordance} fall short of modeling actual object use and manipulation.  In particular, learning to segment specified object parts~\cite{nguyen2016detecting,sawatzky2017weakly,myers2015affordance,nguyen2017object} can capture annotators' expectations of what is important, but is detached from real interactions, which are dynamic, multi-modal, and may only partially overlap with part regions (see \reffig{fig:concept}).
Secondly, existing methods are limited by their heavy supervision and/or sensor requirements.  They  assume access to training images with manually drawn masks or keypoints~\cite{roy2016multi,do2017affordancenet,fang2018demo2vec} and some leverage additional sensors like depth~\cite{koppula2014physically,zhu2016inferring,zhu2015understanding} or force gloves~\cite{castellini2011using}, all of which restrict scalability.  Such bottlenecks also deter generalization: exemplars are often captured in artificial lab tabletop environments~\cite{myers2015affordance,koppula2014physically,sawatzky2017weakly} and labeling cost naturally restricts the scope to a narrow set of objects.

In light of these issues, we propose to learn affordances that are \emph{grounded} in real human behavior directly from videos of people naturally interacting with objects, and without any keypoint or mask supervision.
Specifically, we introduce an approach to infer an object's \emph{interaction hotspots}---the spatial regions most relevant to human-object interactions. Interaction hotspots link \emph{inactive} objects at rest not only to the actions they afford, but also to \emph{how} they afford them.
By learning hotspots directly from video, we sidestep issues stemming from manual annotations, avoid imposing part labels detached from real interactions, and discover exactly how people interact with objects in the wild.

Our approach works as follows. First, we use videos of people performing everyday activities to learn an action recognition model %
that can recognize the array of afforded actions when they are \emph{actively in progress} in novel videos.
Then, we introduce an anticipation model to distill the information from the video model, such that it can estimate how a static image of an \emph{inactive} object transforms during an interaction.  In this way, we learn to anticipate the plausible interactions for an object at rest (\eg, perceiving ``cuttable" on the carrot, despite no hand or knife being in view).
Finally, we propose an activation mapping technique tailored for fine-grained object interactions to derive interaction hotspots from the anticipation model.
Thus, given a new image, we can hypothesize interaction hotspots for an object, even if it is not being actively manipulated.

We validate our model on two diverse video datasets: OPRA~\cite{fang2018demo2vec} and EPIC-Kitchens~\cite{damen2018scaling}, spanning hundreds of object and action categories, with videos from both first and third person viewpoints. Our results show that with just weak action and object labels for training video clips, our interaction hotspots can predict object affordances more accurately than prior weakly supervised approaches, with relative improvements up to 25\%. 
Furthermore, we show that our hotspot maps can anticipate object function for novel object classes that are never seen during training.

\begin{figure*}[htb!]
\centering
\includegraphics[width=\linewidth]{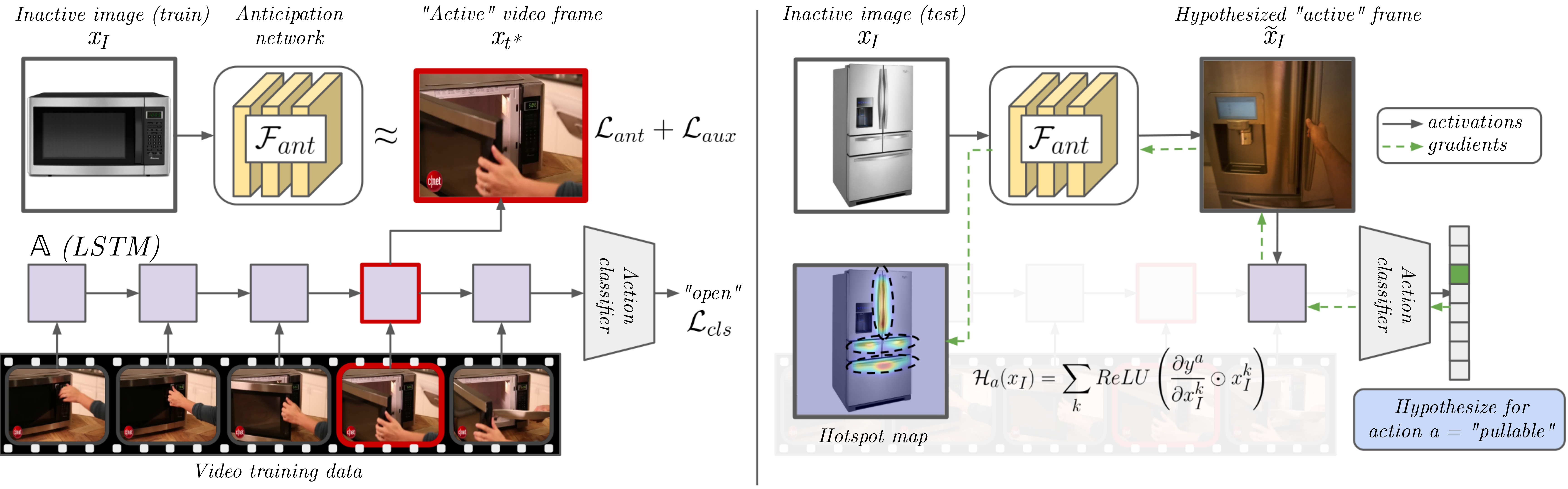}
\caption{\textbf{Illustration of our framework for training (left) and testing (right)}. \textbf{Left panel}: The two components of our model ---the video action classifier and the anticipation module with its associated losses---are jointly trained to predict the action class in a video clip while building an affordance-aware internal representation for objects. \textbf{Right panel}: Once trained, our model generates ``interaction hotspot" maps for a novel \emph{inactive} object image (top left fridge image).  It first hallucinates features that would occur for the object \emph{if it were active} (top right photo), then derives gradient-weighted attention maps over the original image, for each action. %
Our method can infer hotspots even for novel object categories unseen in the training video; for example, learning about opening microwaves helps anticipate how to open the fridge. Note that $x_I$, $\widetilde{x}_I$ are in feature space, not pixel space.
}
\label{fig:model}
\end{figure*}

\section{Approach}

Our goal is to learn ``interaction hotspots": characteristic %
object regions that anticipate and explain human-object interactions, \emph{directly} from video (see \reffig{fig:concept}). In particular, our approach learns to predict afforded actions across a span of objects, then translates the video cues to static images of an object at rest. In this way, without explicit region labels and without direct estimation of physical contact points, we learn to anticipate object use.

\newcommand{\smalltimes}{{\mkern-1.5mu\times\mkern-1.5mu}}
\vspace{0.05in}
\noindent\textbf{Learning Afforded Actions from Video.}
For a video $\mathcal{V}$ with $T$ frames and afforded action class $a$, we encode each frame using a ResNet~\cite{he2016deep} (up to conv5) resulting in features $\{x_1, ..., x_T\}$. These features are then spatially pooled and aggregated over time as follows:
\begin{align}
    g_t(x_t) &= P(x_t) ~~~~~~~~\textrm{for $t=1,\dots,T$}, \\
    h_*(\mathcal{V}) &= \mathbb{A}(g_1,\dots,g_T),
\end{align}
where $P$ denotes the L2-pooling operator and $\mathbb{A}$ is an LSTM~\cite{hochreiter1997long}. The afforded action is then predicted from the aggregated representation using a linear classifier trained with cross entropy $\mathcal{L}_{cls}(h_*, a)$.
Once trained, this model can predict which action classes are observed in a video clip of arbitrary length. 
See \reffig{fig:model} (left) for the architecture. 
\vspace{0.05in}
\noindent\textbf{Anticipation for Inactive Object Affordances.}
This video recognition model alone would focus on ``active" cues directly related to the action being performed (\eg, hands approaching an object), but would not respond strongly to \emph{inactive} instances---static images of objects that are at rest and not being interacted with. In fact, prior work demonstrates that these two incarnations are visually quite different~\cite{deva-adl-2012}.
To account for this, we introduce a distillation-based anticipation module $\mathcal{F}_{ant}$ that transforms the embedding of an inactive object $x_I$, where no interaction is occurring, into its active state where it is being interacted with as $\widetilde{x}_I = \mathcal{F}_{ant}(x_I)$. 
See \reffig{fig:model}, top-left.

During training, the anticipation module is guided by the video action classifier, which selects the appropriate \emph{active state} from a given video as the frame $x_{t^*}$ at which the LSTM is maximally confident of the true action. %
We then define a feature matching loss between (a) the anticipated active state for the inactive object and (b) the active state selected by the classifier network for the training sequence. 
\begin{equation} \label{eqn:anticipation}
   \mathcal{L}_{ant}(x_I, x_{t^*}) = ||P(\widetilde{x}_I) - P(x_{t^*}) ||_2.
\end{equation}
Additionally, we make sure that the newly anticipated representation $\widetilde{x}_I$ is predictive of the afforded action and compatible with our video classifier, using an auxiliary classification loss from a single step of the LSTM $\mathcal{L}_{aux}(h_1(\widetilde{x}_I),a)$.

Overall, these components allow our model to estimate what a static inactive object may potentially look like---in feature space---if it were to be interacted with. They provide a crucial link between classic action recognition and affordance learning.

\vspace{0.05in}
\noindent\textbf{Interaction Hotspot Activation Mapping.}
Finally, we devise an activation mapping approach through $\mathcal{F}_{ant}$ to discover our hotspot maps.
For a particular inactive image embedding $x_I$ and an action $a$, we compute the gradient of the score for the action class with respect to each channel of the embedding. These gradients are used to weight individual spatial activations in each channel, acting as an attention mask over them. The positive components of the resulting tensor are retained and accumulated over all channels in the input embedding to give the final hotspot map for the action: 
\begin{equation}
    \mathcal{H}_a(x_{I}) = \sum_k ReLU\left(\frac{\partial y^a}{\partial  x_I^k} \odot x_I^k\right),
\label{eq:gradcam1}
\end{equation}
where $x_I^k$ is the $k^{th}$ channel of the input frame embedding and $\odot$ is the element-wise multiplication operator. 

We further enhance our hotspots by using dilated, unit stride convolutions in the last two residual stages, increasing our heatmap resolution ($n=28$) to capture finer details. 

In summary, we jointly train our action recognition model and our anticipation model using the combined loss $(\mathcal{L}_{cls} + \mathcal{L}_{ant} + \mathcal{L}_{aux})$ to learn features that can anticipate object use in a video (\reffig{fig:model}, left). Once trained, we generate hotspots, on an inactive test image (\reffig{fig:model}, right), by hypothesizing its \emph{active} interaction embedding $\widetilde{x}_I$, and use it to predict the afforded action scores. Using \refeqn{eq:gradcam1} we generate one heatmap over $x_I$ for each afforded action. This stack of heatmaps are the \emph{interaction hotspots}.

\begin{table*}[t]
\resizebox{2\columnwidth}{!}{
\centering
\begin{tabular}{rl|ccc|ccc|c@{\hskip 0.1in}|ccc|ccc|}
\multicolumn{2}{c}{}                     &          \multicolumn{3}{c}{OPRA}                    &           \multicolumn{3}{c}{EPIC} &\multicolumn{1}{c}{}&        \multicolumn{3}{c}{OPRA}                    &           \multicolumn{3}{c}{EPIC}                   \\
\cmidrule{3-8} \cmidrule{10-15}
&                                        & KLD $\downarrow$ & SIM $\uparrow$ & AUC-J $\uparrow$ & KLD $\downarrow$ & SIM $\uparrow$ & AUC-J $\uparrow$ && KLD $\downarrow$ & SIM $\uparrow$ & AUC-J $\uparrow$ & KLD $\downarrow$ & SIM $\uparrow$ & AUC-J $\uparrow$ \\
\cmidrule{2-8} \cmidrule{10-15}
& \SC{center bias}                       & 11.132           & 0.205          & 0.625            & 10.660           & 0.222          & 0.634            && 6.281            & 0.244          & 0.680            & 5.910            & 0.277          & 0.699            \\
\ldelim[{6}{1mm}[{\rotatebox[origin=c]{90}{WS}}]
& \SC{lstm+grad-cam}                     & 8.573            & 0.209          & 0.620            & 6.470            & 0.257          & 0.626            && 5.405            & 0.259          & 0.644           & 4.508           & 0.255          & 0.664             \\
& \SC{egogaze} \cite{huang2018predicting}& 2.428            & 0.245          & 0.646            & 2.241            & 0.273          & 0.614            && 2.083            & 0.278          & 0.694           & 1.974           & 0.298          & 0.673             \\
& \SC{mlnet} \cite{mlnet2016}            & 4.022            & 0.284          & 0.763            & 6.116            & 0.318          & 0.746            && 2.458            & 0.316          & 0.778           & 3.221           & 0.361          & 0.799             \\     
& \SC{deepgazeII} \cite{deepgaze2016}    & 1.897            & 0.296          & 0.720            & 1.352            & 0.394          & 0.751            && 1.757            & 0.318          & 0.742           & 1.297           & 0.400          & 0.793             \\     
& \SC{salgan} \cite{pan2017salgan}       & 2.116            & 0.309          & 0.769            & 1.508            & 0.395          & 0.774            && 1.698            & 0.337          & 0.790           & 1.296           & \B{0.406}      & 0.808             \\     
& \SC{ours}                              & \B{1.427}        & \B{0.362}      & \B{0.806}        & \B{1.258}        & \B{0.404}      & \B{0.785}        && \B{1.381}        & \B{0.374}      & \B{0.826}       & \B{1.249}       & 0.405          & \B{0.817}         \\     
\cmidrule{2-8} \cmidrule{10-15}

\ldelim[{2}{1mm}[{\rotatebox[origin=c]{90}{SS}}]
& \SC{img2heatmap}                       & 1.473            & 0.355          & 0.821            & 1.400            & 0.359          & 0.794            && 1.431            & 0.362          & 0.820           & 1.466           & 0.353          & 0.770             \\ 
& \SC{demo2vec} \cite{fang2018demo2vec}  & 1.197            & 0.482          & 0.847            & --               & --             & --               && --               & --             & --              & --              & --             & --                \\ 
\cmidrule{2-8} \cmidrule{10-15}
\end{tabular}
}
\vskip -0.1in
\subcaptionbox*{\textbf{Grounded affordance prediction}}[.7\linewidth]{}  \hskip -0.7in
\subcaptionbox*{\textbf{Generalization to novel objects}}[.3\linewidth]{} \vskip -0.1in

\caption{\textbf{Interaction hotspot prediction results on OPRA and EPIC}. \textbf{Left:} Our model outperforms other weakly supervised (WS) methods in all metrics, and approaches the performance of strongly supervised (SS) methods \emph{without} the privilege of heatmap annotations during training. \textbf{Right:} Not only does our model generalize to new \emph{instances}, but it also accurately infers interaction hotspots for novel object \emph{categories} unseen during training. The proposed hotspots generalize on an object-function level. Values are averaged across three splits of object classes. ($\uparrow$/$\downarrow$ indicates higher/lower is better.)  \SC{Demo2Vec}~\cite{fang2018demo2vec} is available only on OPRA and only for seen classes.}
\vspace*{-0.2in}
\label{tbl:hotspot-eval}
\end{table*}

\section{Experiments}  \label{sec:exp}

We evaluate our model on two datasets---\textbf{OPRA}~\cite{fang2018demo2vec}, a product review dataset that comes with videos of people demonstrating product functionalities (\eg, pressing a button on a coffee machine), along with a paired catalog image of the product. The dataset spans 7 actions over $\sim$16k training instances; \textbf{EPIC-Kitchens}~\cite{damen2018scaling}, a large scale egocentric video dataset of people performing daily activities in a kitchen environment. There are $\sim$40k training videos, spanning 352 objects and 125 actions.

Each dataset comes with a set of static, inactive images, labeled with heatmaps for where the interaction takes place, which we use for evaluation.\footnote{We crowd-source annotations for heatmaps on static images from EPIC, resulting in 1.8k annotated instances over 20 action and 31 objects} We stress that (1) the annotated heatmap is used \emph{only} for evaluation, and (2) the ground truth is well-aligned with our objective, since annotators were instructed to watch an interaction video clip to decide what regions to annotate for an object's affordances.

We compare our method to several baselines and existing methods: (1) \SC{Center Bias}: to account for any center bias in our data; (2) \SC{LSTM+Grad-CAM}: Grad-CAM~\cite{selvaraju2017grad} derived heatmaps from a standard LSTM action recognition model; (3) \SC{Saliency}: A set of recent, off-the-shelf models to estimate image saliency including \SC{egogaze}~\cite{huang2018predicting}, \SC{mlnet}~\cite{mlnet2016}, \SC{deepgazeII}~\cite{deepgaze2016} and \SC{salgan}~\cite{pan2017salgan}; (4) \SC{Demo2Vec}~\cite{fang2018demo2vec}: a supervised method trained using heatmaps and videos; (5) \SC{Img2Heatmap}: a simplified supervised model that does not use videos.  We report error as KL-Divergence, following~\cite{fang2018demo2vec}, as well as other metrics (SIM, AUC-J) from the saliency literature~\cite{bylinskii2018different}.

\vspace{0.05in}
\noindent\textbf{Grounded Affordance Prediction}.
\reftbl{tbl:hotspot-eval} (Left) summarizes the results.  Our model outperforms all other weakly-supervised methods in all metrics across both datasets. 
On OPRA, our model achieves relative improvements of up to 25\% (KLD) compared to the strongest baseline, and matches one of the strongly supervised baseline methods on two metrics.
On EPIC, our model achieves relative improvements up to 7\% (KLD).

\begin{figure}[t!]
\centering
\includegraphics[width=\linewidth]{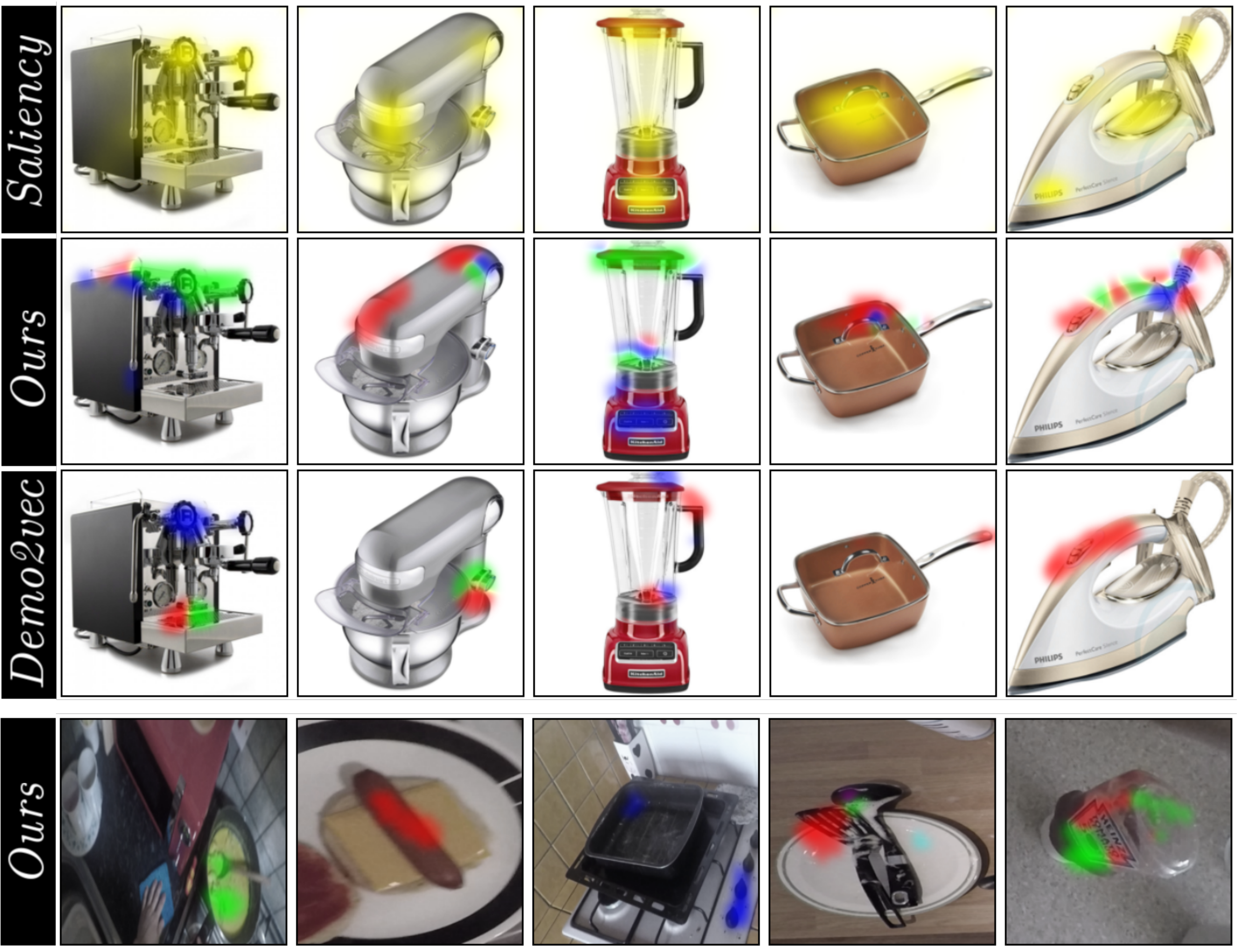}
\vspace*{-0.2in}%
\caption{\textbf{Affordance heatmaps on inactive images.} \textbf{Top:} Predicted affordance heatmaps for \emph{hold}, \emph{rotate}, \emph{push} (red, green, blue) on OPRA. \textbf{Bottom row:} Predicted heatmaps for \emph{cut}, \emph{mix}, \emph{turn-on} (red, green, blue), on EPIC. Our model highlights spatial affordances consistent with how people interact with the objects. Note that \SC{saliency}~\cite{pan2017salgan} produces only a single ``importance" map (yellow).  \textbf{Last column:} failure cases. Best viewed in color.  
} %
\vspace*{-0.1in}
\label{fig:opra-qual}
\end{figure}

The baselines have similar trends across datasets. The \SC{LSTM+Grad-CAM} baseline in \reftbl{tbl:hotspot-eval} demonstrates  that simply training an action recognition model is clearly insufficient to learn affordances. 

All saliency methods perform worse than our model--- they produce a single ``importance" heatmap, which cannot explain objects with multiple affordances. This can be seen in our qualitative results (\reffig{fig:opra-qual}). Our model highlights multiple distinct affordances for an object (\eg, the knobs on the coffee machine as ``rotatable" in column 1) after only watching videos of object interactions, while \SC{Saliency} methods highlight \emph{all} salient object parts in a single map, regardless of the interaction in question. 
\SC{img2heatmap} and \SC{demo2vec} generate better heatmaps, but at the cost of strong supervision. Our method approaches their accuracy without using any manual heatmaps for training.

\begin{figure}
\centering
\includegraphics[width=1\columnwidth]{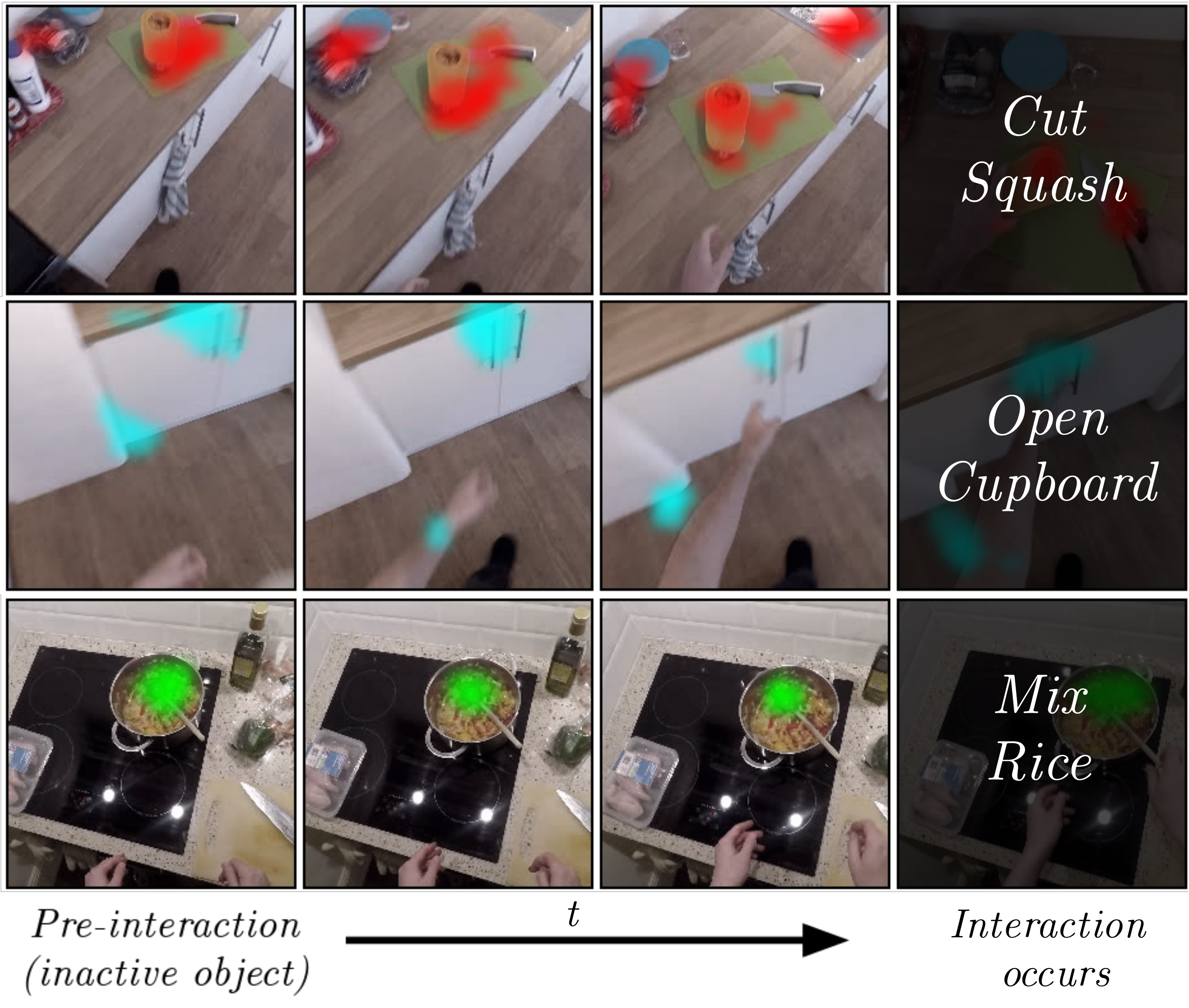}
\vspace*{-0.25in}
\caption{\textbf{Interaction hotspots on EPIC videos of unseen object classes.} %
Our model anticipates interaction hotspots for inactive objects at rest (first column), \emph{even before} the interaction happens. Critically, the object categories shown here were \emph{not} seeing during training; our model learns to generalize interaction hotspots.} 
\label{fig:epic-video}
\vspace*{-0.25in}
\end{figure}

\vspace{0.05in}
\noindent\textbf{Generalization to Novel Objects.} Can interaction hotspots infer how \emph{novel} object categories work?
We divide our object categories into disjoints sets of familiar and unfamiliar objects, and only train on video clips with familiar objects. We test if our model can successfully infer heatmaps for novel, unfamiliar objects, implying that a general sense of object \emph{function} is learned that is not strongly tied to object \emph{identity}. 
\reftbl{tbl:hotspot-eval} (Right) shows mostly similar trends as the previous section.  On OPRA, our model outperforms all baselines in all metrics, and is able to infer the hotspot maps for unfamiliar object categories, despite never seeing them during training. 
On EPIC, our method remains the best weakly supervised method. \reffig{fig:epic-video} illustrates that our model---which was never trained on some objects (\eg, cupboard, squash)---is able to anticipate characteristic spatial locations of interactions even \emph{before} the interaction occurs.

{\small
\bibliographystyle{ieee}
\bibliography{egbib}
}

\end{document}